\documentclass[conference, 11pt]{IEEEtran}

\IEEEoverridecommandlockouts
\usepackage{cite}
\usepackage{amsmath,amssymb,amsfonts}
\usepackage{algorithmic}
\usepackage{graphicx}
\usepackage{textcomp}
\usepackage{xcolor}
\usepackage{amsthm}
\usepackage{url}

\def\BibTeX{{\rm B\kern-.05em{\sc i\kern-.025em b}\kern-.08em
    T\kern-.1667em\lower.7ex\hbox{E}\kern-.125emX}}

\newcommand{\DefinedAs}{\triangleq}
\newcommand{\Transpose}{^{\mathsf{T}}}

\newcommand{\Input}{\mathbf{x}}
\newcommand{\Key}{\mathbf{k}}
\newcommand{\Query}{\mathbf{q}}
\newcommand{\Value}{\mathbf{v}}

\newcommand{\Rotation}{\mathbf{R}}
\newcommand{\RotationAngle}{\theta}

\newcommand{\AttentionWeight}{\mathbf{A}}

\newcommand{\Time}{t}
\newcommand{\OtherTime}{u}

\newcommand{\KeyTransformation}{\mathbf{K}}
\newcommand{\QueryTransformation}{\mathbf{Q}}
\usepackage{comment}
\usepackage{multirow}
\usepackage{booktabs}
\begin{document}

\title{Benchmarking Rotary
Position Embeddings for Automatic Speech Recognition
}
\author{\IEEEauthorblockN{1\textsuperscript{st} Shucong Zhang}
\IEEEauthorblockA{\textit{AI Center, Samsung} \\
Cambridge, UK \\
s1.zhang@samsung.com}
\and
\IEEEauthorblockN{2\textsuperscript{nd} Titouan Parcollet}
\IEEEauthorblockA{\textit{AI Center, Samsung} \\
Cambridge, UK \\
t.parcollet@samsung.com}
\and
\IEEEauthorblockN{3\textsuperscript{rd} Rogier van Dalen}
\IEEEauthorblockA{\textit{AI Center, Samsung} \\
Cambridge, UK \\
r.vandalen@samsung.com}
\and
\IEEEauthorblockN{4\textsuperscript{th} Sourav Bhattacharya}
\IEEEauthorblockA{\textit{AI Center, Samsung} \\
Cambridge, UK \\
sourav.b1@samsung.com}
}

\maketitle

\begin{abstract}
Self-attention relies on positional embeddings to encode input order. Relative Position (RelPos) embeddings are widely used in Automatic Speech Recognition (ASR). However, RelPos has quadratic time complexity to input length and is often incompatible with fast GPU implementations of attention. In contrast, Rotary Positional Embedding (RoPE) rotates each input vector based on its absolute position, taking linear time to sequence length, implicitly encoding relative distances through self-attention dot products. Thus, it is usually compatible with efficient attention. However, its use in ASR remains underexplored. This work evaluates RoPE across diverse ASR tasks with training data ranging from 100 to 50,000 hours, covering various speech types (read, spontaneous, clean, noisy) and different accents in both streaming and non-streaming settings. ASR error rates are similar or better than RelPos, while training time is reduced by up to 21\%. Code is available via the SpeechBrain toolkit.
\end{abstract}


\section{Introduction}
Transformer \cite{vaswani2017attention} models have demonstrated impressive results in natural language processing (NLP) and speech processing. At the heart of these models lies the self-attention mechanism \cite{vaswani2017attention}, which does not inherently capture the positional order of input vectors; consequently, any permutation of these vectors yields identical outputs. To address this issue, absolute positional embeddings \cite{vaswani2017attention} were introduced. This method uses a sinusoidal function that takes the positions of the input sequence to generate positional embeddings, which are then added to each input vector at the corresponding position.

However, absolute positional embeddings do not explicitly model the relative positions between input vectors, which may lead to poor generalization \cite{dai2019transformer}. 
To address this limitation, Relative positional embeddings (RelPos) \cite{dai2019transformer} have been proposed. 
Compared to absolute positional embeddings, RelPos has shown superior performance in NLP and speech processing tasks.
Currently, widely used speech processing toolkits, including SpeechBrain \cite{ravanelli2021speechbrain, ravanelli2024open}, ESPnet \cite{watanabe2018espnet}, and NeMo \cite{kuchaiev2019nemo}, use RelPos as their default positional embedding method. 

To encode relative distance information, RelPos introduces  dot-product operations between input vectors and positional embeddings, 
in addition to the dot-products computed between input vectors in the original self-attention mechanism. 
These additional dot-products lead to a quadratic time complexity with respect to sequence length for encoding positional information.
Furthermore, RelPos is not compatible with many fast GPU implementations of self-attention, such as FlashAttention \cite{dao2022flashattention} and PyTorch \cite{paszke2019pytorch} GPU-accelerated attentions, since these implementations in general require positional embeddings to be integrated into the input vectors.

To address these issues, in the NLP community, Rotary Position Embedding (RoPE) \cite{su2024roformer} has been proposed and widely adopted as an alternative to RelPos. It has demonstrated superior performance compared to RelPos in NLP tasks and has been employed in state-of-the-art NLP models, such as the Llama family \cite{touvron2023llama}. 
RoPE rotates each input vector by an angle proportional to its absolute position. 
Then, the outputs of the dot product between two rotated vectors in the self-attention mechanism depends solely on the original unrotated vectors and the angular difference between the rotated vectors. 
Since this angular difference is determined by the relative positional difference of the input vectors, the self-attention mechanism effectively encodes relative distance. 
Rotating each input vector takes a linear time with respect to the input sequence length, and thus encoding the positional information through RoPE has a linear time-complexity.
Further, since it does not alter any operation inside the self-attention mechanism, RoPE is in general compatible with GPU-accelerated implementations of attention.



However, RoPE has not yet been well studied for automatic speech recognition (ASR) tasks. Though \cite{li2021conformer, samarakoon2022conformer} compared RoPE with RelPos for ASR, 
these studies were limited to only one or two  tasks. \cite{jeffries2024moonshine} employed RoPE to build a large-scale ASR system, but they did not provide a comparison between RelPos and RoPE, nor did they explore the benefits of RoPE.

In this work, we conduct a comprehensive evaluation of RoPE against RelPos across diverse ASR tasks. Our experiments include reading and spontaneous speech, clean and noisy environments, native as well as accented speech, and different languages, with training data ranging from approximately 100 hours to 50,000 hours.
We evaluate both online streaming and offline models. 
Across all experiments, RoPE achieves lower or similar ASR word error rates compared to RelPos with up to 21\% training time reduction.
Further, to the best of our knowledge, we note that widely used open-source speech processing toolkits currently lack either RoPE implementation or recipes for utilising RoPE in ASR tasks. 
To address this gap, we release our RoPE implementation \footnote{\url{https://github.com/speechbrain/speechbrain/blob/develop/speechbrain/nnet/attention.py}} along with training recipes \footnote{\url{https://github.com/speechbrain/speechbrain/tree/develop/recipes/LibriSpeech}
} for all experiments through SpeechBrain 1.0, enabling the research community to conduct further investigations. 

\section{Rotary Positional Embeddings}

The self-attention mechanism transforms the input sequence $\Input_{1:T} \in \mathbb{R}^{d \times T}$, which consists of $T$ vectors of dimension $d$, into a query sequence $\Query_{1:T}$, a key sequence $\Key_{1:T}$, and a value sequence $\Value_{1:T}$. Then, dot-products are applied between each query vector and key vector pair. The normalised dot product yields $\AttentionWeight \in \mathbb{R}^{T \times T}$, which is referred to as the attention weights. The output of the self-attention is a weighted sum of the value sequence, where the weights are given by 
$\AttentionWeight$. With these operations, the outputs of self-attention are invariant to the order of the input vectors if no positional embeddings are used.

To address this, rotary positional embeddings (RoPE) use transformations based on the absolute position in the sequence that yield a relative positional embedding. The following explains how this paradox is resolved. The key idea is to use an orthonormal matrix, i.e., a rotation matrix, which explains the “rotary” part of the embeddings’ name. This matrix is raised to the power corresponding to the absolute position of each key or query vector and is then used to transform the vector in the corresponding position. 

We examine how RoPE transforms absolute positional rotations into relative encoding in a two-dimensional setting, as this setting best illustrates the two key aspects of this process.
First, if the rotation matrix describes a rotation of angle $\RotationAngle$, then the $\Time$th power of this matrix performs a rotation of angle $\Time\cdot\RotationAngle$. So each time step is rotated an additional $\RotationAngle$ compared with its predecessor. Second, in the dot product used to compare the key and query, a transpose is used implicitly. The transpose of an orthonormal matrix is the inverse of the matrix. In two dimensions, the inverse matrix rotates the vector back. In the dot product between each key and each query, the part of the rotations related to their absolute positions cancel out, and the part related to their relative positions remains.

Mathematically, the transformations can be described as follows. In vanilla self-attention without any positional embedding, the key vector $\Key_{\Time}$ and query vector $\Query_{\Time}$ are formed by transforming $\Input_{\Time}$ with two linear transformations:
\begin{align}
    \Key_{\Time} &= \KeyTransformation \cdot \Input_{\Time}
    ;&
    \Query_{\Time} &= \QueryTransformation \cdot \Input_{\Time}
    .
\end{align} where $\QueryTransformation \in \mathbb{R}^{d \times d} $ and $\KeyTransformation \in \mathbb{R}^{d \times d} $.
The attention weights are then derived from the dot products between the keys and queries as follows:
\begin{align}
    \log \AttentionWeight_{\Time\OtherTime} &=
    \Key_{\Time}\Transpose \cdot \Query_{\OtherTime}.
    \label{eq:attention_weight:standard}
\intertext{
Instead, for rotary (as well as for absolute) positional embeddings, the key and query are first further transformed, so the dot product will be between $\Key{_\Time}'$ and queries $\Query_{\Time}'$:
}
    \log \AttentionWeight_{\Time\OtherTime} &=
    {\Key'_{\Time}}\Transpose \cdot \Query'_{\OtherTime}.
    \label{eq:attention_weight:transformed}
\end{align}
The transformation used for rotational positional embeddings are given by $\Rotation_{\Time}$, a matrix specific to the time step:
\begin{align}
    \Key'_{\Time} &\DefinedAs \Rotation_{\Time} \Key_{\Time}
    ;&
    \Query'_{\OtherTime} &\DefinedAs \Rotation_{\OtherTime} \Query_{\OtherTime}
    \label{eq:key_query:transformed}
\end{align}
To achieve this transformation, we begin by choosing $\Rotation_{1}$ as a square, orthonormal matrix.
Thus, all columns of $\Rotation_{1}$ are orthogonal to each other and have a Euclidean length of $1$, so it can be interpreted as a rotation matrix.
Next, we define the transformation for time step $\Time$ as the $\Time$th power of $\Rotation_1$: \begin{align} \Rotation_{\Time} &\DefinedAs \Rotation_1^{\Time}. \label{eq:rotation:exponentiation} \end{align} For a two-dimensional matrix, this results in a rotation by an angle of $\Time \cdot \RotationAngle$. Since $\Rotation_{\Time}$ is orthonormal, the transformation preserves the Euclidean norm of the vector. It will become important that since $\Rotation_{\Time}$ is orthonormal, its transpose is its inverse:
\begin{align}
    \Rotation_{\Time}\Transpose &\DefinedAs \Rotation_{\Time}^{-1}.
    \label{eq:rotation:transpose}
\end{align}
For a two-dimensional rotation matrix, this means that the transpose of a matrix rotates it back.
In general:
\begin{align}
    \Rotation_{\Time} \Rotation_{\OtherTime} &= \Rotation_{\Time + \OtherTime}
    ; &
    \Rotation_{\Time}^{-1} \Rotation_{\OtherTime} &= \Rotation_{\OtherTime-\Time}.
    \label{eq:rotation:relative}
\end{align}

The effect of the transformations can be seen by combining the previous equations. By substituting \eqref{eq:key_query:transformed} into \eqref{eq:attention_weight:transformed} and then applying \eqref{eq:rotation:relative}, we obtain: 
\begin{align}
    \log \AttentionWeight_{\Time\OtherTime} &=
        (\Rotation_{\Time} \Key_{\Time})\Transpose \cdot \Rotation_{\OtherTime} \Query_{\OtherTime}
        \notag\\&
        =
        \Key_{\Time}\Transpose \Rotation_{\Time}^{-1} \Rotation_{\OtherTime} \Query_{\OtherTime}
        =
        \Key_{\Time}\Transpose \Rotation_{\OtherTime-\Time} \Query_{\OtherTime}
    \label{eq:attention_weight:rotation:relative}
    .
\end{align} 
Thus, compared to \eqref{eq:attention_weight:standard}, \eqref{eq:attention_weight:rotation:relative} can be interpreted as incorporating a rotation matrix $\Rotation_{\OtherTime-\Time}$ into the dot-product computation. Importantly, the effective rotation depends solely on $\OtherTime-\Time$, the relative distance between the key and query positions. Consequently, the resulting attention weight is independent of the absolute positions of the keys and queries, thereby implementing a relative positional embedding.

Following \cite{su2024roformer} we now present the explicit formulas for the the rotation matrix  $\mathbf{R}_t$ which employs two-dimensional rotations with a small angle $\RotationAngle$.:
\begin{align}
    \mathbf{R}_t &=  \begin{bmatrix}
   \mathbf{R}_{t1} & \mathbf{0} & \dots & \mathbf{0} \\ 
   \mathbf{0}& \mathbf{R}_{t2} & \dots  & \mathbf{0}\\ 
  \vdots & \vdots  &  \ddots & \vdots \\ 
   \mathbf{0}& \mathbf{0} &  \dots  & \mathbf{R}_{t{d\over2}} 
 \end{bmatrix} \\
    \mathbf{R}_{ti} &=  \begin{bmatrix}
   \cos t\theta_i &   -\sin t\theta_i \\ 
    \sin t\theta_i  & \cos t\theta_i
 \end{bmatrix}
\end{align} where the rotation angle for each pair of consecutive dimension is defined as $\RotationAngle_i = 10000 ^{-2(i-1)\over d}, i \in [1,2,\dots, {d\over 2}]$.
Due to the sparsity of the rotation matrix, the operation of rotating any vector  $\mathbf{y} \in \mathbb{R}^d$ using $\mathbf{R}_t$ can be reformulated for computational efficiency as:
\begin{align}
\mathbf{R}_t \mathbf{y} = 
\begin{bmatrix}
 y_1\\
 y_2\\
 y_3\\
 y_4\\
 \vdots\\
 y_{d-1}\\
 y_d\\
\end{bmatrix} \circ 
\begin{bmatrix}
 \cos t\theta_1 \\
 \cos t\theta_1\\
 \cos t\theta_2\\
 \cos t\theta_2\\
 \vdots\\
 \cos t\theta_{d\over 2}\\
 \cos t\theta_{d\over 2}\\ 
\end{bmatrix} + 
\begin{bmatrix}
 y_2\\
 y_1\\
 y_4\\
 y_3\\
 \vdots\\
 y_d\\
 y_{d-1}\\
\end{bmatrix} \circ 
\begin{bmatrix}
 -\sin t\theta_1 \\
 \sin t\theta_1\\
 -\sin t\theta_2\\
 \sin t\theta_2\\
 \vdots\\
 -\sin t\theta_{d\over 2}\\
 \sin t\theta_{d\over 2}\\ 
\end{bmatrix} 
\label{eq:rope_efficient_compute}
\end{align}

Applying RoPE to the query and key sequences
through \eqref{eq:rope_efficient_compute} leads to a computational cost of $O(T)$.
Furthermore, since RoPE preserves the original operations of the self-attention mechanism, it remains broadly compatible with modifications to self-attention, such as efficient attention variants. In contrast, Relative Position Encoding (RelPos) alters the dot product computation, which introduces a computational cost of $O(T^2)$ for encoding positional information. The modifications to the dot product computation also make RelPos incompatible with fast GPU attention implementations.

\section{Experiments}
We first compare the training speed of rotational and relative positional embeddings across varying utterance lengths under controlled experimental conditions, highlighting RoPE’s linear time complexity versus RelPos’s quartic time complexity. We then benchmark RelPos and RoPE with common model architectures across multiple datasets, evaluating their training speed and automatic speech recognition (ASR) performance through word error rates (WERs).

\subsection{Experimental Protocol}
In all the experiments, we use the Conformer encoder \cite{gulati2020conformer}, since Conformer encoders give impressive ASR results and are widely used in the speech processing community.
We conduct experiments to measure training speed for various input lengths, as well as to evaluate actual training speed and speech recognition performance on different datasets for both offline ASR and online streaming ASR.
For measuring the training speed across different input lengths, we use a Conformer CTC  \cite{graves2006connectionist} model, minimizing computational overhead unrelated to the encoder itself.
For offline ASR tasks, we employ Conformer encoder-decoder architectures with joint CTC-attention training \cite{kim2017joint}, pure CTC models, and Conformer-Transducer models \cite{graves2012sequence}. Online streaming ASR experiments use Conformer-Transducer architectures exclusively.
Complete details of the model configurations and training recipes are provided in the released official SpeechBrain recipes.

We compare Conformer encoders with either relative or rotational positional embeddings.
For experiments aimed at isolating the effects of positional encoding, we integrate RoPE and RelPos into the same SpeechBrain 1.0 self-attention implementation.

In addition, to leverage RoPE’s compatibility with fast GPU implementations of self-attention, we further implement Conformer encoders combining RoPE with PyTorch’s GPU-accelerated self-attention implementation. To be more specific, we use the PyTorch C++ math scaled dot product attention. FlashAttention is not considered since each of our training batches may contain sequences with different length. 
In contrast, RelPos is incompatible with such optimisations due to its requirement of modifying the dot-product operation within the self-attention mechanism.

We emphasise that our comparison of RoPE and RelPos is restricted to the Conformer encoder, with no evaluation for the decoders with RoPE, as this work specifically investigates RoPE’s effectiveness for modeling acoustic inputs. 

The original paper proposing RoPE uses a value of 10,000 for the parameter $\theta$ \cite{su2024roformer}. Recent studies have suggested that increasing the $\theta$ value can enhance the performance of large language models \cite{xiong2024effective} and large speech language models \cite{nguyen2025spirit}. This is because a larger $\theta$ encourages higher attention scores for distant tokens. Compared to the default $\theta$ value, a larger value thereby potentially improves the ability of self-attention components to capture long-range dependencies. However, in our Conformer ASR experiments, we observed larger $\theta$ does not consistently yield better speech recognition results; in fact, it sometimes degrades performance. One plausible explanation is that, for Transformer- and Conformer-based speech recognition models, previous research has demonstrated the importance of local information within self-attention mechanisms \cite{zhang2021usefulness, shim2021understanding, parcollet2024summarymixing}. Consequently, larger $\theta$ values, which bias self-attention toward long-range dependencies, may not necessarily lead to improved ASR performance. Therefore, throughout all experiments reported in this paper, we maintain the default $\theta$ value of 10,000 as recommended by the original paper \cite{su2024roformer}.



\subsection{Training Speed of Different Input Lengths}
We measure the training speed of Conformer CTC models across different input lengths with controlled experiments.
In this task, the Conformer encoder has 12 layers with 512 dimensionality. 
With this configuration, the encoder with RelPos has 76M trainable parameters, which is slightly more than the encoder with RoPE at 73M, because RelPos has trainable parameters while RoPE has none. 
We randomly create input sequences with length from 1 to 50 seconds with a 16 kHZ sample rate.
There are 5 random output tokens for each second of input.
The outputs have a vocabulary size of 5,000.
We use the PyTorch Profiler to measure the GPU time of one forward-backward pass of processing the input sequence with each length.
For each sequence length, we repeat the experiment 500 times and report the average GPU time.
An isolated NVIDIA A40 GPU is used.


\begin{figure}[t]
    \centering
    \includegraphics[width=\columnwidth]{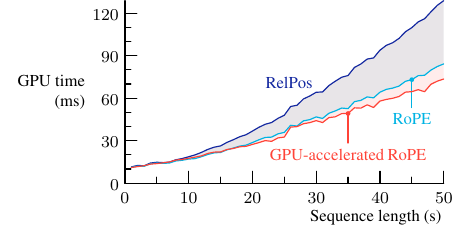}


    \includegraphics[width=\columnwidth]{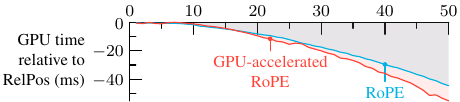}

    \vspace{1mm}

    \caption{The GPU time for Conformer encoders with RelPos, RoPE, and RoPE combined with PyTorch GPU-accelerated attention, as well as the relative GPU times of RoPE-based Conformer encoders compared to RelPos-based Conformer encoders across different sequence lengths.}
    \label{fig:train speed}
\end{figure}

Figure \ref{fig:train speed} shows the GPU time across different input lengths for Conformer encoder with RelPos, RoPE, and RoPE combined with PyTorch GPU-accelerated attention, respectively.
For all the models, the GPU time increases quadratically with input length, which is the expected behavior of self-attention based models.
However, compared to RoPE models, the training time of the RelPos model increases much faster with increasing input lengths, suggesting that RelPos has a higher time complexity relative to RoPE.
Figure \ref{fig:train speed} further illustrates the relative GPU times between RoPE models and the RelPos model. These differences increase quadratically with input length, which confirms RoPE’s linear time complexity against RelPos’s quartic time complexity.
While the PyTorch GPU-accelerated attention further enhances the training speed of the RoPE model, the gain is less dramatic compared to the speed gain achieved by replacing RelPos with RoPE within the same self-attention implementation.
RelPos is not compatible with the GPU-accelerated attention.

\subsection{Encoder-decoder Experiments}

We equip the encoder of the Conformer encoder-decoder models with RoPE and RelPos. For this set of set of experiments, we have used the joint CTC-attention architecture.  
We first consider the LibriSpeech \cite{panayotov2015librispeech} dataset, a long-standing, widely used benchmark. This dataset contains 960 hours of training data of reading speech. For LibriSpeech, the encoder-decoder models consist of 12 encoder layers, 6 encoder layers, and 512-dimensional hidden vectors. The model with RelPos or RoPE has 109M or 106M trainable parameters, respectively. All models are trained on 4 NVIDIA A40 GPUs with a 0.5 hour batch size.


\begin{table}[t!]
\caption{Speech Recognition results of Conformer encoder-decoder models with different positional embedding on the LibriSpeech and the Libriheavy datasets. ``Efficient attention'' refers to the PyTorch GPU-accelerated attention.}
\centering
\begin{tabular}{l@{~~}c@{~~~}c@{~~~}c@{~~~~}c}
\toprule
\multirow{2}{*}{\textbf{\begin{tabular}[l]{@{}l@{}}Positional \\ Embedding\end{tabular}}} & \multicolumn{3}{c}{\textbf{Word Error Rates}}   & \multirow{2}{*}{\textbf{\begin{tabular}[c]{@{}c@{}}Train \\ Time \end{tabular}}} \\
  & \begin{tabular}[c]{@{}c@{}}dev-\\ clean\end{tabular} & \begin{tabular}[c]{@{}c@{}}test-\\ clean\end{tabular} & \begin{tabular}[c]{@{}c@{}}test-\\ other\end{tabular} &   \begin{tabular}[c]{@{}c@{}} \\ relative\end{tabular}   \\ \midrule
\textbf{LibriSpeech 960 h}  &  &   &   &   \\
RelPOS & 1.88  & 2.00  & 4.57 & 1.00x  \\
RoPE  & \textbf{1.85}  & \textbf{1.96}  & \textbf{4.50}  & 0.86x  \\ 
RoPE, efficient attention  & 1.88  & 1.97  & \textbf{4.50}  & 0.81x  \\ \hline
\textbf{Libriheavy 50,000 h}  &  &    &     &    \\
RelPOS  & 1.58   & 1.74    & 3.92 & 1.00x   \\
RoPE & \textbf{1.57}  & \textbf{1.68}  & \textbf{3.69} &  0.83x  \\ 
RoPE, efficient attention & 1.58  & 1.70  & 3.71 &  0.79x  \\ \bottomrule
\end{tabular}
\label{tab:libri}
\end{table}

Table~\ref{tab:libri} shows the ASR results for models on LibriSpeech. With the same self-attention implementation, RoPE consistently outperforms the RelPos in terms of WER, with a 14\% training time reduction.
When RoPE combined with PyTorch GPU-accelerated attention, WERs increase slightly but remain lower than the WERs of the RelPos model. The training time reduction further rises from 14\% to 19\%, highlighting the efficiency advantages of RoPE over RelPos.

To evaluate the performance of RelPos and RoPE with scaled-up training data, we consider the Libriheavy \cite{kang2024libriheavy} dataset. The Libriheavy dataset consists of 50,000 hours of reading speech. The training set of Libriheavy contains the training set of LibriSpeech, making it an extended version of LibriSpeech. We train models with 14 encoder layers, 6 decoder layers, and a hidden dimension 640, resulting in 165M trainable parameters for the model with RelPos, and 160M trainable parameters for the models with RoPE. Training is conducted on 8 NVIDIA A100 GPUs with an one-hour batch size. 
Table ~\ref{tab:libri} shows the models trained with the Libriheavy dataset demonstrate significantly lower WERs across all the evaluation sets compared to their LibriSpeech trained counterparts, reflecting the impact of the substantially increased training data. Notably, RoPE maintains its ASR performance advantage over RelPos in this large-scale setting. 
When using the same self-attention implementation as RelPos, RoPE gives 17\% training time reduction.
With PyTorch GPU-accelerated attention, the RoPE model further saves 21\% training time.

Since LibriSpeech and Libriheavy consist of reading speech, we further examine the performance of RoPE in a more challenging ASR scenarios. We use the Loquacious Set \cite{parcollet2025loquacious}, which contains 25,000 hours of diverse speech data. The dataset includes hundreds of thousands of speakers with varied accents, covering read speech, spontaneous speech, and talks under diverse acoustic conditions (clean, noisy, and far-field). For this dataset, all the experimental configurations are identical to the configurations of the Libriheavy experiment.

\begin{table}[t!]
\caption{Speech Recognition results of Conformer encoder-decoder models with different positional embedding on the 25,000 hours of the Loquacious Set. ``Efficient attention'' refers to the PyTorch GPU-accelerated attention.}
\label{tab:loquacious}
\centering
\begin{tabular}{lccc}
\toprule
\multirow{2}{*}{\textbf{\begin{tabular}[l]{@{}l@{}}Positional \\ Embedding\end{tabular}}} & \multicolumn{2}{c}{\textbf{Word Error Rates}}   & \multirow{2}{*}{\textbf{\begin{tabular}[c]{@{}c@{}}Train \\ Time \end{tabular}}} \\
  & \begin{tabular}[c]{@{}c@{}}dev \end{tabular} & \begin{tabular}[c]{@{}c@{}}test \end{tabular} & \begin{tabular}[c]{@{}c@{}} \\ relative\end{tabular}   \\ \midrule
RelPOS     &   8.11   &  8.97   &  1.00x  \\ 
RoPE     &   8.07   &  8.98  &  0.88x  \\ 
RoPE, efficient attention      &  \textbf{8.05}   &  \textbf{8.92}   &  0.82x  \\ \hline
\end{tabular}
\end{table}

Table~\ref{tab:loquacious} shows the experimental results for the Loquacious Set. When using the same self-attention implementation, the RoPE model achieves a 0.04 absolute WER reduction on the dev set compared to the RelPos model, with a marginal 0.01 absolute WER increase on the test set, while reducing training time by 12\%. When integrated with PyTorch GPU-accelerated attention, the RoPE model delivers the best dev and test WERs and achieves an 18\% reduction in training time. These experiments demonstrate that RoPE provides equivalent or superior ASR performance on large-scale, real-world diverse data while substantially improving training speed.

\begin{table}[t]
\caption{Speech Recognition results of Conformer encoder-decoder models with different positional embedding on the CommonVoice 18.0 dataset.}
\centering
\label{tab:cv}
\begin{tabular}{lccc}
\toprule
\multirow{2}{*}{\textbf{\begin{tabular}[l]{@{}l@{}}Positional \\ Embedding\end{tabular}}} & \multicolumn{3}{c}{\textbf{Word Error Rates}}  \\  & \begin{tabular}[c]{@{}c@{}}\textbf{Dutch}\\ 111 h\end{tabular} & \begin{tabular}[c]{@{}c@{}}\textbf{Italian} \\ 357 h\end{tabular} & \begin{tabular}[c]{@{}c@{}}\textbf{French}\\ 1025 h\end{tabular} \\ \hline
RelPOS     & 33.27    & 9.17    & 8.50 \\
RoPE  & \textbf{33.14}  & \textbf{9.09}  & \textbf{8.46}\\ \bottomrule
\end{tabular}
\end{table}

To evaluate the ASR performance of RoPE for non-English languages, we conduct experiments using the CommonVoice 18.0 \cite{ardila2019common} dataset. Table~\ref{tab:cv} shows that RoPE achieves superior ASR results for Dutch, Italian, and French, with the amount of training data ranging from about 100 to 1,000 hours. These experiments demonstrate RoPE's effectiveness across varying amounts of training data and multiple languages. No training speed improvement is observed for RoPE in this setting, as the dataset primarily consists of short utterances (e.g., $<5$ seconds). The linear-complexity RoPE does not achieve faster training speeds than the quadratic-complexity RelPos for such short sequences. However, as shown in previous experiments, RoPE’s speed advantage increases as sequence lengths grow longer. Due to our limited computation resource, we do not conduct GPU-accelerated attention experiments. 

\subsection{Conformer CTC}
The previous subsection presented the experimental results of the joint CTC-attention model. This subsection provides experimental results from pure CTC models. We compare the speech recognition performance of Conformer CTC models utilising RoPE and RelPos on the LibriSpeech dataset. The models are configured with 18 encoder layers and a hidden dimension of 256, resulting in approximately 28.8M  trainable parameters for the model employing RelPos and approximately 27.6M parameters for the model with RoPE. The batch size used is equivalent to 3.8 hours of audio.

\begin{table}[t!]
\caption{Speech Recognition results of Conformer CTC models with different positional embedding on the LibriSpeech. ``Efficient attention'' refers to the PyTorch GPU-accelerated attention.}
\label{tab:libri-ctc}
\centering
\begin{tabular}{l@{~~}c@{~~~}c@{~~~}c@{~~~~}c}
\toprule
\multirow{2}{*}{\textbf{\begin{tabular}[l]{@{}l@{}}Positional \\ Embedding\end{tabular}}} & \multicolumn{3}{c}{\textbf{Word Error Rates}}   & \multirow{2}{*}{\textbf{\begin{tabular}[c]{@{}c@{}}Train \\ Time \end{tabular}}} \\
  & \begin{tabular}[c]{@{}c@{}}dev-\\ clean\end{tabular} & \begin{tabular}[c]{@{}c@{}}test-\\ clean\end{tabular} & \begin{tabular}[c]{@{}c@{}}test-\\ other\end{tabular} &   \begin{tabular}[c]{@{}c@{}} \\ relative\end{tabular}   \\ \midrule
RelPOS & \textbf{3.81}  & \textbf{4.03}  & \textbf{11.06} & 1.00x  \\
RoPE, efficient attention  & 3.94  & 4.07  & 11.07  & 0.81x  \\ \bottomrule
\end{tabular}
\end{table}

Table~\ref{tab:libri-ctc} summarises the speech recognition results. In these experiments, RelPos marginally outperforms RoPE, showing relative improvements ranging from 0.001 to 0.06. Nevertheless, the Word Error Rates (WERs) of both models remain the same level. Notably, RoPE reduces training time by approximately 19\% compared to RelPos. We hypothesize that the small WER gap could be bridged through hyper-parameter optimisation, which we reserve for future investigation.

\subsection{Conformer Transducer}


Finally, we evaluate the Conformer Transducer equipped with RoPE and RelPos. 
We use the Voxpopuli \cite{wang2021voxpopuli} dataset, which comprises 550 hours of European Parliament debate recordings. 
The models consist a 12-layer Conformer encoder and a one layer LSTM predictor, and 512-dimensional hidden vectors.
This leads to 79M trainable parameters for the model with RelPos and 76M trainable parameters for the model with RoPE.
The model is trained with dynamic chunk training \cite{yao21_interspeech}, enabling the same model to process both full utterances for offline decoding and variable chunk sizes of frames for online streaming decoding. 

\begin{table}[t!]
\caption{Results for streaming speech recognition of Conformer Transducer models with different positional embeddings on the Voxpopuli dataset. Offline decoding uses full context; streaming uses chunks of 1280ms, 640ms and 320ms.}
\label{tab:vox}
\centering
\begin{tabular}{lccccc}
\toprule
\multirow{2}{*}{\begin{tabular}[c]{@{}l@{}}\textbf{Positional}\\ \textbf{Embedding}\end{tabular}} & \multicolumn{4}{c}{\textbf{Word Error Rate}}  & \multirow{2}{*}{\textbf{\begin{tabular}[c]{@{}c@{}}Train \\ Time \end{tabular}}} \\
   & \multicolumn{4}{c}{ \textbf{Decoding Chunk size (ms)}}   \\
   & offline    & 1024    & 640    & 320 & relative  \\ \hline
RelPOS  & 9.28     & 10.02  & 10.36 & 11.13 & 1.00x \\
RoPE    & \textbf{8.91}  & \textbf{9.75} & \textbf{10.27}  & \textbf{11.06}  & 0.81x \\ \bottomrule                                                              
\end{tabular}
\end{table}

Table~\ref{tab:vox} shows that the RoPE model has 19\% training time reduction compared to the RelPos model. For the evaluation, with the offline decoding setting, RoPE gives lower WERs compared to the RelPos model. Then, for the online streaming evaluations across varying chunk sizes, RoPE also consistently achieves lower WERs compared to the RelPos baseline. This set of experiments has demonstrated the effectiveness of RoPE for both of offline and online streaming ASR scenarios. 
\section{Conclusion}
 In this work, we benchmark Rotary Position Embedding (RoPE) for automatic speech recognition using Conformer encoder-decoder, Conformer CTC, and Conformer Transducer models. Experiments encompass diverse training data volumes (ranging from 100 to 50,000 hours), various speech types (read, spontaneous, clean, noisy), different accents, and multiple languages (English, French, Italian, and Dutch).
 The results demonstrate that Conformer models equipped with RoPE consistently gives similar or better results compared to those using RelPOS across both streaming and non-streaming scenarios with up to 21\% training time reduction, indicating RoPE's broad effectiveness for ASR tasks. To facilitate further research in this area, we release our implementation and training recipes with SpeechBrain.

\bibliographystyle{IEEEtran}
\bibliography{mybib}

\end{document}